\documentclass{article}
\usepackage{amsmath}
\usepackage{xcolor}
\usepackage[normalem]{ulem}
\usepackage{graphicx,float,booktabs,tabularx,amsmath,amssymb}

\usepackage[final,nonatbib]{tackling_climate_workshop_style}

\usepackage[utf8]{inputenc} % allow utf-8 input
\usepackage[T1]{fontenc}    % use 8-bit T1 fonts
\usepackage{hyperref}       % hyperlinks
\hypersetup{hidelinks,
  pdftitle={Learning When to Transfer: Reinforcement Learning for Carbon-Footprint Modeling in New Manufacturing Domains},
  pdfauthor={Muhammad Umar Farooq, Daniel Cooper}}
\usepackage{url}            % simple URL typesetting
\usepackage{booktabs}       % professional-quality tables
\usepackage{amsfonts}       % blackboard math symbols
\usepackage{graphicx}       % for including graphics
\usepackage{nicefrac}       % compact symbols for 1/2, etc.
\usepackage[protrusion=true,expansion=false]{microtype} % microtypography

\pdfmapfile{=ps2pk35.map}
\pdfmapfile{+cm.map}
\pdfmapfile{+cmextra.map}
\pdfmapfile{+symbols.map}

\title{Learning When to Transfer: Reinforcement Learning for Carbon-Footprint Modeling in New Manufacturing Domains}

\author{Muhammad Umar Farooq \qquad Daniel R. Cooper \\
Mechanical Engineering Department, University of Michigan \\
2350 Hayward Street, Ann Arbor, 48109, MI, United States \\
Correspondence: \texttt{muf@umich.edu}}

\begin{document}

\maketitle

\begin{abstract}
Manufacturers often need to characterize a new material, process, or product before enough operating data are available to train reliable predictive models for this new, target domain. Existing manufacturing operations contain potentially useful knowledge to accelerate this characterization, but indiscriminate reuse of data from existing source domains can cause negative transfer in learning of the new, target domain when the manufacturing domains differ sufficiently. This work proposes a reinforcement-learning method for selective knowledge transfer, where its reward scores each opportunity to transfer against supervised alternatives, so negative transfer is avoided and controlled during forward prediction instead of being diagnosed afterwards. Given limited data from a new target domain, the learned policy chooses among predictors based on individual source operations, subsets of the sources, or the supervised alternatives: fully pooled sources, or target data alone. A validation gate deploys the transferred prediction only when it improves upon the supervised alternatives. The method is designed for learning manufacturing variables such as process energy and material consumption, which populate life-cycle inventories and support carbon accounting, as well as performance and quality outcomes. We demonstrate the approach by quantifying the embodied greenhouse-gas emissions of parts produced via material-extrusion additive manufacturing, using data from 729 experiments spanning five materials: ABS, PETG, TPU, PLA, and recycled PLA. The response combines measured fabrication electricity with polymer-specific cradle-to-gate production factors, while 22 material, process, and toolpath features characterize each experiment. In leave-one-material-out evaluation, the proposed method achieves a mean test RMSE of 5.2 g.CO$_2$e per part, 20.0\% lower than target-only training  (32.5\% lower when only 5--10\% of the target training data are available) and 10.3\% lower than indiscriminate source-target pooling. The validation gate reduces the observed negative-transfer rate from 33\% to 17\%, compared with 49\% for pooling. These results demonstrate how manufacturers can selectively reuse data from existing operations to accelerate carbon-footprint modeling for a newly introduced material while limiting harmful transfer.
\end{abstract}

\section{Introduction}

Manufacturing and production activities account for approximately one-fifth of global carbon emissions \cite{wef_carbon_2022}. Manufacturing decarbonization depends on the timely development of accurate life-cycle inventories (LCIs), which underpin cradle-to-gate life-cycle assessments (LCAs) and enable informed, environmentally motivated decision-making. Decisions about a new material, process, or product therefore require foreground data on energy requirements, material consumption, yield, and the generation of wastes and by-products. Yet these measurements are local to the operating domain, and compiling domain-specific LCIs remains costly and time-consuming \cite{farooq_data-driven_2026}.

Mechanistic simulations and measurements from existing operations can reduce this burden. Previous work has already shown how correlation-aware multi-fidelity learning can combine low-fidelity physics-based priors with a small number of experiments (providing sparse high-fidelity data) to accelerate development of manufacturing LCIs \cite{farooq_multifidelity_2026}. Historical operating data create a complementary opportunity: predictors already fitted to prior domains can encode useful process--response structure. However, a new domain may alter material properties, feasible settings, toolpaths, or dominant energy mechanisms. Pooling all historical observations can then introduce bias, while choosing a source by descriptive similarity alone may overlook response-level compatibility \cite{wang2019characterizing}. Therefore, using data and/or models from existing manufacturing domains to accelerate LCI model development on a new domain requires explicit decisions on \emph{whether}, \emph{what}, and \emph{how much} of this existing knowledge to transfer.

This issue is visible in additive manufacturing (AM). Data-driven LCA and G-code models estimate energy and environmental performance earlier in design \cite{naser_automating_2023,yu_code-based_2024}, but reported models are commonly tied to one material, machine, geometry, or sensing configuration. AM datasets are often small because experiments are slow and costly \cite{fullington_leveraging_2024}; transfer learning can reduce this burden, including cross-material quality monitoring \cite{li_transfer_2024}. More generally, manufacturing studies use source screening, instance reweighting, representation alignment, and multi-source ensembles to improve transfer \cite{mao_tensor_2024,qiang2024multi,gao2024innovative}. These methods establish that source usefulness is heterogeneous, but typically optimize within a predetermined transfer pipeline. They do not treat target-only learning and alternative transfer configurations as competing deployment choices.

This work addresses the gap by framing low-shot manufacturing prediction as a contextual, one-step reinforcement-learning problem. One policy, trained across source domains, weights individual-source, source-subsets, fully pooled sources, and target-only predictors. A separate validation gate deploys its mixture only when it improves on the supervised (target-only or fully pooled) baseline alternatives. Reinforcement learning is used because the output is a decision: the correct transfer choice for a new material is never labelled, its value is defined relative to a supervised alternative, and the policy replaces that alternative only when held-out evidence favours it~\cite{thomas2015high,laroche2019safe}. The contributions are: (i) a material-agnostic decision formulation that represents domain distance, source competence, predictor disagreement, and target-data budget; (ii) a controlled leave-one-material-out protocol that separates fitting, model-selection, and test evidence; and (iii) an empirical analysis of accuracy and negative transfer across five material (polymer AM) domains and seven target-data budgets, against direct learning (the supervised target-only alternative) and policy ablations.

\section{Problem formulation and method}
\begin{figure}[!htbp]
  \centering
  \includegraphics[width=\linewidth]{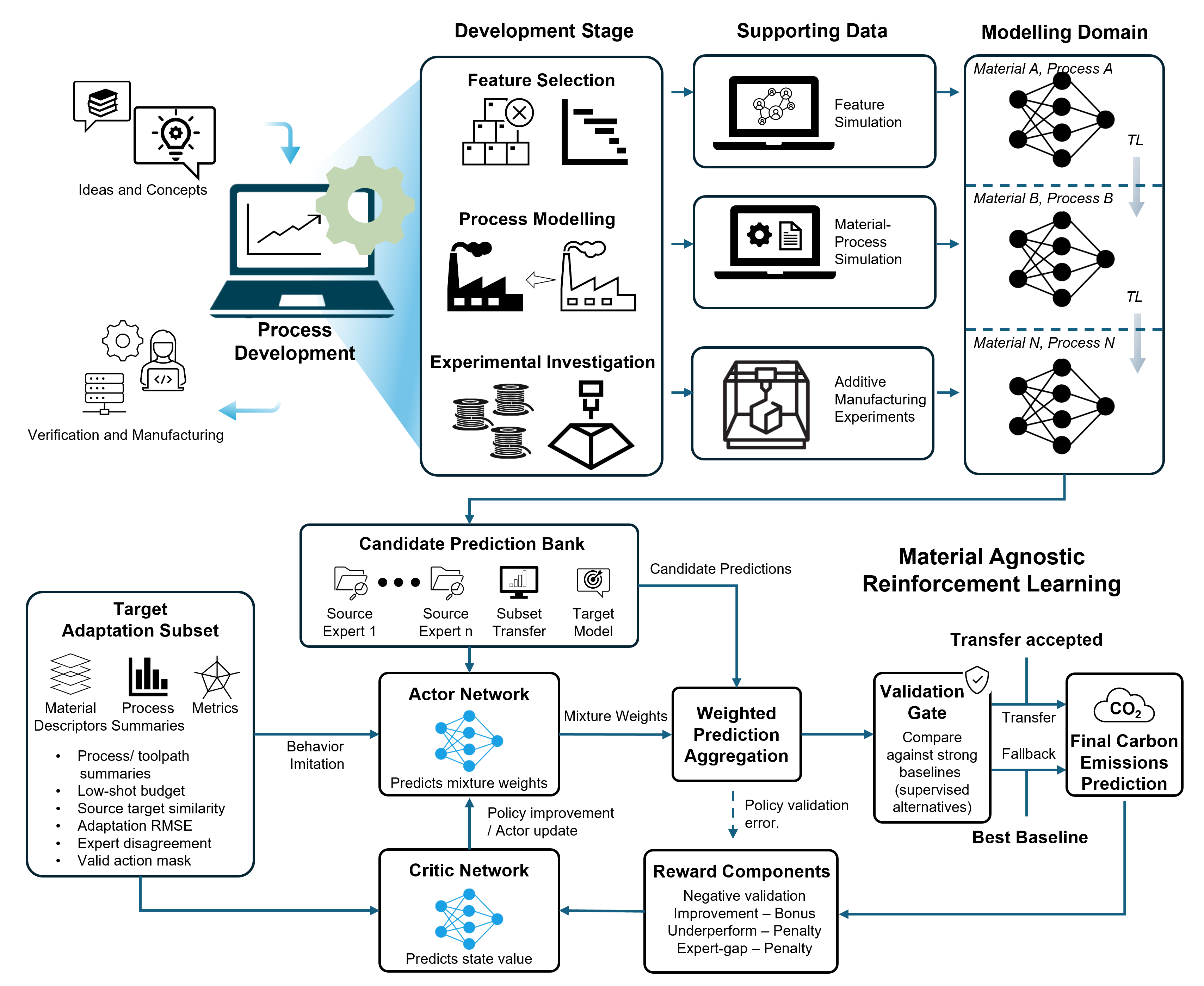}
  \caption{Manufacturing-development context and transfer-policy workflow. Top: conceptual links between process development, supporting data, and manufacturing domains. Bottom: target-domain evidence informs an actor that assigns weights to candidate predictors. A critic supports reward-based policy training, and a validation gate selects between the policy mixture and the stronger supervised alternative.}
  \label{app:fig:workflow}
\end{figure}

Figure~\ref{app:fig:workflow} illustrates the manufacturing-domain policy workflow, with its key components summarized below and described in greater detail in the Appendix.

\subsection{Experimentation and carbon inventory}

\begin{figure}[htbp]
  \centering
  \includegraphics[width=\linewidth]{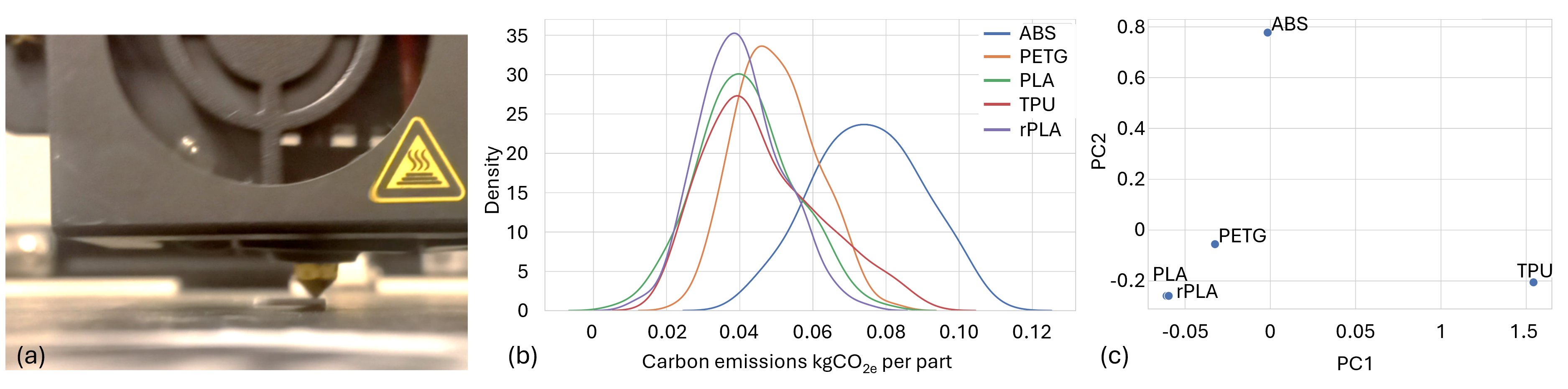}
  \caption{Experimental setup and characterization of the five material domains:(a) material-extrusion printing; (b) estimated density curves of the per-part carbon response defined in Eq.~\ref{eq:carbon}, in $\mathrm{kgCO_2e/part}$; and (c) material descriptors projected onto the first two principal components, PC1 and PC2.}
  \label{fig:demonstration}
\end{figure}

The dataset contains 729 controlled material-extrusion prints: 145 each for ABS, PETG, PLA, and recycled PLA (rPLA), and 149 for TPU. All trials use the same specimen geometry and harmonized schema. For each material, a design of experiments is conducted over the AM process settings. Electrical power is recorded at 1~Hz; process settings and G-code are archived. The 22 predictors comprise 11 material descriptors, six controllable process parameters, and five toolpath aggregates. For experiment $i$ using polymer $j$, the per-part response (in $\mathrm{kgCO_2e}$) is
\begin{equation}
C_i=C_EE_i+C_{m,j}m_i,
\label{eq:carbon}
\end{equation}
where $E_i$ is the measured fabrication electricity (kWh), $m_i$ is the filament mass used (kg), $C_E=0.587~\mathrm{kgCO_2e/kWh}$ is the location-specific electricity factor \cite{noauthor_emission_nodate}, and $C_{m,j}$ ($\mathrm{kgCO_2e/kg}$) is a polymer-specific cradle-to-gate filament production factor \cite{koci_introducing_2022,council_cradle--gate_2022,material_plastic_nodate,corbion_environmental_nodate}. Equation~\ref{eq:carbon} is an operational-plus-material inventory indicator, not a complete comparative LCA. Figure~\ref{fig:demonstration} shows the printing setup, the resulting carbon-response distributions, and a two-dimensional projection of the material descriptors.

\subsection{Material-agnostic transfer policy}

For target domain $t$ and fitting budget $b$, the candidate bank $\mathcal C_t(b)=\{g_1,\ldots,g_C\}$ contains predictors trained on each source separately, selected source subsets plus target data, all sources plus target data, and target data alone. The policy state records $b$, target-set summaries, fold-standardized material distances, source-expert fitting errors, prediction disagreement, and validation performance of the target-only and pooled supervised models. Material-agnostic means that one decision rule is shared across materials; material differences remain explicit state variables.

An action $a\in\Delta^{C-1}$ assigns nonnegative weights summing to one to the candidate predictors, giving $g_a(x)=\sum_{j=1}^{C}a_jg_j(x)$. Let $\widehat a_\theta(s)$ denote the weights produced by the policy for deployment in state $s$; the policy predictor is $g_{\mathrm{pol}}(x)=g_{\widehat a_\theta(s)}(x)$. Behaviour cloning initializes the actor toward validation-optimal convex mixtures. Actor--critic refinement rewards low validation normalized RMSE (NRMSE) and improvement over the stronger supervised alternative, while penalizing underperformance relative to that alternative and the validation-optimal mixture. Each target--budget case has one state, one model-selection action, and one validation reward: this is contextual reinforcement learning, not sequential process control.

\subsection{Evaluation and validation gate}

Policy generalization is evaluated using leave-one-material-out: four materials form policy-training cases and the fifth is the unseen target. Each material is partitioned before transfer into mutually exclusive training (50\%), validation (20\%), and test (30\%) sets. Target fitting uses 5, 10, 20, 40, 60, 80, or 100\% of the training partition, with eleven repetitions per budget; below 100\% each repetition draws a different fitting subset, and at 100\% the repetitions share the full training partition and differ only in learner random seeds. Policy-training rewards use the validation data of source materials treated as pseudo-targets. For the held-out target, validation data provide the supervised model-selection evidence, validation-state features, and deployment-gate comparison, without updating the policy parameters; thus, the budgets quantify scarce \emph{fitting} data, not total labelled data. The untouched target test set measures generalization after all choices are fixed.

At deployment, the baseline model $g_{\mathrm{fb}}$ is defined by selecting the better-performing model (i.e., the one with lower target-validation NRMSE) among the target-only model and the model trained on all sources plus the target fitting subset. The complete policy (actor--critic + validation gate) deploys
\begin{equation}
g_{\mathrm{final}} =
\begin{cases}
g_{\mathrm{pol}},
& \mathrm{NRMSE}_{\mathrm{pol,val}}
  \leq \mathrm{NRMSE}_{\mathrm{fb,val}},\\
g_{\mathrm{fb}}, & \text{otherwise}.
\end{cases}
\label{eq:fallback}
\end{equation}
This validation-based deployment heuristic does not guarantee lower test error.

\section{Results and discussion}

\subsection{Transfer structure and overall performance}
\label{sec:transfer-results}

Source-expert transfer is directional: PETG and TPU transfer effectively in both directions; however, PETG transfers effectively to rPLA but not vice versa. Overall, ABS is the most difficult target. Material distance (the Wasserstein distance between standardized material-descriptor distributions) summarizes differences in material descriptors. For fixed empirical distributions, this distance is symmetric, whereas predictive transfer errors can differ between directions. Moreover, similar material descriptors do not imply similar relationships between process inputs and carbon response. The policy therefore uses source-expert errors on the target fitting subset, disagreement among source-expert predictions for the same inputs, and target-data availability alongside distance.

Across held-out materials, fitting budgets, and subset draws, the complete policy ("actor-critic + gate" in Table 1) achieves the lowest mean test RMSE, $5.2~\mathrm{gCO_2e/part}$ (Table~\ref{tab:results}). This is 20.0\% lower than target-only fitting and 10.3\% lower than fully pooled model. The relative benefit of the complete policy grows as target data become scarce. The ablations separate the actor-critic policy learning from the effect of the validation gate: raw actor--critic obtains $5.6~\mathrm{gCO_2e/part}$, behaviour cloning with the gate obtains $5.3~\mathrm{gCO_2e/part}$, and actor--critic with the gate obtains $5.2~\mathrm{gCO_2e/part}$. The ordering indicates that the central advantage comes from learning over a diverse candidate bank while retaining a strong supervised alternative. Under the gate, actor--critic refinement lowers the negative-transfer rate from 28\% to 17\% (McNemar $p=4\times10^{-5}$ across repetitions for these materials), while mean RMSE reduces by only 0.1~g$CO_2e/part$.

\begin{table}[htbp]
\centering
\caption{Held-out test RMSE ($\mathrm{kgCO_2e/part}$) and negative-transfer rate by target fitting budget $b$. Negative-transfer rate is 100 times the number of cases in which a method's test RMSE exceeds target-only test RMSE, divided by the number of evaluated cases. Each budget contains $5$ materials $\times$ $11$ repetitions $=55$ cases against each budget; overall results contain $5\times7\times11=385$ cases. RMSE entries are means; the Avg.\ RMSE and Test $R^2$ entries report mean $\pm$ standard deviation across all cases. AC: actor--critic; BC: behaviour cloning; Pooled: all sources + target.}
\label{tab:results}
\scriptsize
\setlength{\tabcolsep}{2.6pt}
\begin{tabular}{@{}l|ccccc|cccc@{}}
\toprule
 & \multicolumn{5}{c|}{Test RMSE ($\mathrm{kgCO_2e/part}$)} & \multicolumn{4}{c}{Negative-transfer rate (\%)} \\
$b$ & AC + gate & BC + gate & Raw AC & Pooled & Target only & AC + gate & BC + gate & Raw AC & Pooled \\
\midrule
0.05 & \textbf{0.0072} & 0.0076 & 0.0076 & 0.0075 & 0.0113 & \textbf{4} & 11 & 11 & 15 \\
0.10 & \textbf{0.0065} & 0.0067 & 0.0069 & 0.0069 & 0.0090 & \textbf{2} & 16 & 22 & 20 \\
0.20 & \textbf{0.0058} & 0.0059 & 0.0061 & 0.0062 & 0.0069 & \textbf{18} & 29 & 33 & 38 \\
0.40 & \textbf{0.0047} & 0.0049 & 0.0051 & 0.0054 & 0.0053 & \textbf{22} & 33 & 38 & 51 \\
0.60 & \textbf{0.0044} & 0.0044 & 0.0047 & 0.0053 & 0.0046 & \textbf{27} & 31 & 42 & 69 \\
0.80 & \textbf{0.0039} & 0.0041 & 0.0043 & 0.0048 & 0.0042 & \textbf{25} & 33 & 45 & 65 \\
1.00 & \textbf{0.0037} & 0.0039 & 0.0043 & 0.0047 & 0.0040 & \textbf{24} & 40 & 42 & 82 \\
\midrule
Avg. & \textbf{0.0052}$\pm$0.0021 & 0.0053$\pm$0.0022 & 0.0056$\pm$0.0022 & 0.0058$\pm$0.0023 & 0.0065$\pm$0.0033 & \textbf{17} & 28 & 33 & 49 \\
Test $R^2$ & \textbf{0.776}$\pm$0.215 & 0.726$\pm$0.213 & 0.720$\pm$0.189 & 0.663$\pm$0.265 & 0.493$\pm$0.661 & & & & \\
\bottomrule
\end{tabular}
\end{table}

\subsection{Target-data budget and source composition}
\label{sec:budget-results}

Relative predictive performance varies with both the target
material and the target fitting-data budget
(Table~\ref{tab:results}; Fig.~\ref{app:fig:budgets}). When target data are scarce ($b\leq0.10$), the complete policy lowers RMSE by 32.5\% relative to target-only fitting, with negative-transfer rates of 2--4\%, against 15--20\% for pooling. TPU shows the clearest transition: target-only fitting is weak at a target-data budget of 5--10\% but becomes preferable after approximately 40\% of the target training partition. Across all 35 material--budget cells, the complete policy results in mean RMSE on the target test set within 15\% of the best fixed rule chosen in hindsight; whereas, behaviour cloning with the gate, and the raw actor--critic are up to 86\% and 134\% higher, respectively.

Useful information is also compositional rather than purely pairwise. Frequently, competitive subsets include TPU+rPLA, PETG+TPU, and ABS+PLA. A source that is weak in isolation can still contribute after joint training with other sources and target data. For example, ABS appears in useful multi-source subsets despite being a poor stand-alone source for several targets. Selective subset transfer can therefore exploit complementary structure that is lost under both nearest-source selection (the source with the smallest Wasserstein distance to the target) by material distance and indiscriminate pooling.

\subsection{Validation-gated deployment and manufacturing relevance}
\label{sec:deployment-results}

The validation gate converts this target-dependent transfer structure into a deployment decision. Raw actor--critic exceeds target-only test RMSE in 33\% of evaluated cases; the gate reduces the observed rate to 17\%, compared with 49\% for all-sources-plus-target pooling (Appendix Fig.~\ref{app:fig:deployment}).

The manufacturing relevance of this work is that it enables manufacturers to develop more accurate LCAs of emerging manufacturing domains—including materials, processes, and products—by modeling the response in Eq.~\ref{eq:carbon}. These improved assessments support better-informed environmental decision-making early in development, when design and process choices can have the greatest influence on long-term environmental impacts. Ultimately, this approach helps identify and evaluate pathways toward deep decarbonization.

\section{Conclusion}

Across 729 material-extrusion experiments, the new material-agnostic policy lowers mean test RMSE relative to target-only fitting and indiscriminate pooling, while its validation gate approximately halves the raw policy's observed negative-transfer rate. The results indicate that selective reuse of existing manufacturing data supports accurate and rapid carbon-footprint modeling when manufacturers introduce a new material, with the largest gains when target data are scarce.

\clearpage
{\small
\bibliographystyle{unsrt}
\bibliography{references}
}

\clearpage
\appendix

\section{Experimental protocol and evaluation}
\label{app:protocol}

\subsection{Domains, partitions, and target-data budgets}

Each polymer is treated as a distinct manufacturing domain. The dataset contains 145 experiments each for ABS, PETG, PLA, and recycled PLA (rPLA), and 149 experiments for TPU. Before any transfer experiment, every material is divided into mutually exclusive training (50\%), validation (20\%), and test (30\%) partitions. Source experts and target-side candidates are fitted using training data; validation data are used for model selection, policy rewards, and the deployment gate; the test partition is accessed only after all model and policy decisions are fixed.

For target material $t$, the fitting subset at budget $b$ is
\begin{equation}
  \mathcal A_t(b)\subset\mathcal S^{(t)}_{\mathrm{train}},
  \qquad
  \mathcal B=\{0.05,0.10,0.20,0.40,0.60,0.80,1.00\}.
  \label{app:eq:budget}
\end{equation}
Only $\mathcal A_t(b)$ is used to fit target-dependent candidates. The full 20\% validation partition is retained at every budget and is not counted in $b$. Consequently, the experiment measures scarcity of \emph{target fitting data}, rather than scarcity of all labelled target observations. Table \ref{app:tab:labels} makes this distinction explicit.

\begin{table}[htbp]
\centering
\caption{Approximate target fitting-set sizes. Ranges reflect the difference
between the 145-sample and 149-sample material domains and integer rounding.}
\label{app:tab:labels}
\small
\setlength{\tabcolsep}{5pt}
\begin{tabular}{@{}lrrrrrrr@{}}
\toprule
Budget $b$ & 0.05 & 0.10 & 0.20 & 0.40 & 0.60 & 0.80 & 1.00 \\
\midrule
Target fitting observations & 4 & 7--8 & 14--15 & 29--30 & 43--45 & 58--60 & 72--75 \\
Fixed validation observations & \multicolumn{7}{c}{approximately 29--30 at every budget} \\
\bottomrule
\end{tabular}
\end{table}

Eleven repetitions are used at each budget, giving 385 held-out cases. At $b<1$, the repetitions differ in which training observations form $\mathcal A_t(b)$. At $b=1$, $\mathcal A_t(b)$ is the full training partition, so the eleven repetitions share identical fitting data and differ only in the random seeds of the stochastic tree learners (bootstrap and feature subsampling), whose effect propagates through the candidate predictions to the policy state, action, and gate. Variation at $b=1$ therefore reflects model randomness rather than data sampling and is correspondingly small: the within-material standard deviation of the proposed method across repetitions is 0.08--0.24~g per part at $b=1$, against 0.18--2.0~g at $b=0.05$. In the leave-one-material-out evaluation, four materials generate policy-training cases and the fifth material is held out as an unseen target. Each case is defined by a target material, a fitting-data budget, and a repetition. Reported aggregate metrics combine target materials, budgets, and repetitions.

\subsection{Fold-specific preprocessing and material distance}

Material distance is recalculated within every outer fold. For material descriptor $f$ in fold $k$,
\begin{equation}
  \widetilde{x}^{\,k}_f=
  \frac{x_f-\mu^{\mathrm{meta}}_{f,k}}
       {\sigma^{\mathrm{meta}}_{f,k}},
  \label{app:eq:standardize}
\end{equation}
where $\mu^{\mathrm{meta}}_{f,k}$ and $\sigma^{\mathrm{meta}}_{f,k}$ are fitted
using only the policy-training materials; zero-variance features receive unit
scale. The dimensionless material distance is
\begin{equation}
  d_{\mathrm{mat}}^{(k)}(a,b)
  =\frac{1}{|F_{\mathrm{mat}}|}
   \sum_{f\in F_{\mathrm{mat}}}
   W_1\!\left(\widetilde P^{(f)}_{a,k},
               \widetilde P^{(f)}_{b,k}\right).
  \label{app:eq:distance}
\end{equation}
Material descriptors are constant within a material domain, so the feature-wise Wasserstein distance reduces to the absolute difference between standardized descriptor values. The same fold-specific distance is used in the policy state, the similarity-weighted baseline, and nearest-source pretrain--finetune. State features are transformed using Yeo--Johnson and standardization operations fitted only on the policy-training cases of the outer fold.

\section{Candidate models and comparison methods}
\label{app:candidates}

\subsection{Source experts and candidate prediction bank}

The source-expert model family is

\begin{equation}
  \mathcal M_{\mathrm{tree}}=\left\{\text{Random Forest},\ \text{Extra Trees},\ \text{Gradient Boosting},\ \text{AdaBoost},\ \text{XGBoost}\right\}.
\end{equation}
For each source $s$, the expert is the model with the lowest source-validation RMSE:
\begin{equation}
m_s^\star =
\arg\min_{m\in\mathcal M_{\mathrm{tree}}}
\mathrm{RMSE}\!\left(
y_{\mathrm{val}}^{(s)},
\widehat m_{s,m}(X_{\mathrm{val}}^{(s)})
\right),
\qquad
f_s(x)=\widehat m_{s,m_s^\star}(x).
\label{app:eq:expert}
\end{equation}
Source experts use process-control and toolpath predictors. Material properties are domain constants and are therefore used as domain descriptors, rather than as sample-varying inputs to a within-material source expert.

For target $t$ and budget $b$, the candidate bank $\mathcal C_t(b)=\{g_1,\ldots,g_C\}$ contains: (i) each single-source expert; (ii) supervised models retrained on selected source subsets of size at least two together with $\mathcal A_t(b)$; (iii) target-only scratch; and (iv) all sources plus target. Test predictions from these candidates are stored but are not used during candidate construction, policy training, or deployment selection.

\subsection{Baseline definitions}

Table~\ref{app:tab:baselines} defines every comparison method. Each method uses the fitting and selection data specified below; no method uses the held-out test partition for fitting or selection.

\begin{table}[htbp]
\centering
\caption{Evaluated baselines and the aspect of transfer tested by each method.}
\label{app:tab:baselines}
\scriptsize
\setlength{\tabcolsep}{3.2pt}
\begin{tabularx}{\linewidth}{@{}p{0.19\linewidth}p{0.13\linewidth}X p{0.27\linewidth}@{}}
\toprule
Method & Transfer basis & Rule & Purpose \\
\midrule
Uniform average & Prediction & Equal-weight average of all source experts & Tests whether naive source ensembling is sufficient. \\
Similarity-weighted & Prediction & Inverse weighting by $d_{\mathrm{mat}}^{(k)}$ & Tests whether material proximity alone identifies useful sources. \\
Best single source & Prediction & Source with lowest error on $\mathcal A_t(b)$ & Hard source selection using limited target fitting evidence. \\
Adaptation simplex & Prediction & Nonnegative convex stacking of source predictions fitted on $\mathcal A_t(b)$ & Strong source-only prediction-combination baseline. \\
Target-only scratch & Data & Best tree model fitted only on $\mathcal A_t(b)$ & Non-transfer reference. \\
All sources + target & Data & Best tree model fitted on all source training data plus $\mathcal A_t(b)$ & Indiscriminate supervised transfer reference. \\
Nearest-source pretrain--finetune & Parameters & Pretrain an MLP on the nearest source and fine-tune on $\mathcal A_t(b)$ & Parameter-transfer baseline using the same distance as the policy. \\
Pooled pretrain--finetune & Parameters & Pretrain the same MLP on pooled sources and fine-tune on $\mathcal A_t(b)$ & Parameter-transfer baseline using all source data. \\
Behaviour cloning + gate & Model selection & Imitate validation-optimal candidate weights, then apply the gate & Isolates supervised policy initialization without actor--critic refinement. \\
Raw actor--critic & Model selection & Refine candidate weights using the one-step reward, without the gate & Isolates reinforcement-learning refinement. \\
Actor--critic + gate & Selection + fallback & Raw policy followed by comparison with the stronger supervised reference & Proposed deployed method. \\
\bottomrule
\end{tabularx}
\end{table}

The four prediction-level source baselines can be written
\begin{equation}
  \widehat y(X)=\sum_{s\ne t}w_s f_s(X),\qquad
  w_s\ge0,\quad\sum_{s\ne t}w_s=1.
\end{equation}
Uniform averaging uses $w_s=1/S$, where $S$ is the number of source domains. Similarity weighting uses
\begin{equation}
  w_s=\frac{\max\{d_{\mathrm{mat}}^{(k)}(t,s),10^{-8}\}^{-1}}
  {\sum_{j\ne t}\max\{d_{\mathrm{mat}}^{(k)}(t,j),10^{-8}\}^{-1}}.
\end{equation}
Best-single-source chooses the source with the lowest RMSE on the target fitting
subset. Adaptation simplex solves
\begin{equation}
  w^\star=\arg\min_{w\in\Delta^{S-1}}
  \left\|y_A-\widehat Y_Aw\right\|_2^2,
\end{equation}
where $\widehat Y_A$ stacks source-expert predictions on $\mathcal A_t(b)$.
The neural pretrain--finetune baselines use the full material, process, and
toolpath vector after fold-specific Yeo--Johnson preprocessing, Huber loss,
early stopping on target validation data, and gradient clipping.

\section{Material-agnostic transfer policy}
\label{app:policy}

\subsection{State, action, and behaviour-cloning initialization}

The state contains: (i) the target fitting-data budget; (ii) means and standard deviations of the target fitting subset over the material, process, and toolpath feature groups; (iii) fold-specific standardized source--target material distances; (iv) source-expert errors on the fitting subset; (v) disagreement among source-expert predictions; and (vi) validation performance of the target-only and all-sources-plus-target references.

The action is a vector $a\in\Delta^{C-1}$ of nonnegative candidate weights, producing
\begin{equation}
  g_a(x)=\sum_{j=1}^{C}a_jg_j(x).
  \label{app:eq:mixture}
\end{equation}
An action mask gives unavailable candidates exactly zero probability and zero mixture weight. For available-candidate set $\mathcal V_t$,
\begin{equation}
  p_{\theta,j}(s)=
  \frac{\mathbb I(j\in\mathcal V_t)\exp(z_{\theta,j}(s))}
  {\sum_{\ell\in\mathcal V_t}\exp(z_{\theta,\ell}(s))}.
  \label{app:eq:mask}
\end{equation}

The behaviour-cloning target is the validation-optimal convex mixture
\begin{equation}
  a^\star=\arg\min_{a\in\Delta^{C-1},\;a_j=0\ \forall j\notin\mathcal V_t}
  \left\|y_{\mathrm{val}}-P_{\mathrm{val}}a\right\|_2^2,
  \label{app:eq:teacher}
\end{equation}
and the actor is initialized with
\begin{equation}
  \mathcal L_{\mathrm{BC}}=-\sum_{j\in\mathcal V_t}a_j^\star
  \log p_{\theta,j}(s).
  \label{app:eq:bc}
\end{equation}
Here $s$ is the state of a target--budget case, $p_{\theta,j}(s)$ are the masked, normalized candidate weights of Eq.~\ref{app:eq:mask}, and $\pi_\theta(a\mid s)$ in Appendix~\ref{app:policy}.2 denotes the Dirichlet density over the continuous weight vector $a$; both are computed from the same actor network. Because each case is a single decision, no time index is needed.

\subsection{One-step actor--critic refinement}

The actor and critic are two-hidden-layer multilayer perceptrons. Positive masked actor outputs parameterize a Dirichlet distribution over candidate weights, and a non-reparameterized action $a\sim\mathrm{Dirichlet}(\alpha_\theta(s))$ is sampled during refinement. The critic predicts the scalar state value. Because every target--budget case contains one state, one model-selection action, and one validation reward, the method is contextual one-step reinforcement learning rather than sequential process control.

For validation prediction $\widehat y$, normalized RMSE is
\begin{equation}
  \mathrm{NRMSE}(\widehat y)=
  \frac{\sqrt{n_{\mathrm{val}}^{-1}
  \sum_i(y_i-\widehat y_i)^2}}
  {\max_i y_i-\min_i y_i}.
  \label{app:eq:nrmse}
\end{equation}
Let $g_{\mathrm{ref}}$ be the better of target-only scratch and all sources plus
target on validation data, and let $g_{\mathrm{best}}$ be the
validation-optimal available mixture (best expert weights). The reward is (explained in Table~\ref{tab:reward_components}):
\begin{equation}
\begin{aligned}
r={}&-\mathrm{NRMSE}_{\mathrm{pol}}
+1.5[\mathrm{NRMSE}_{\mathrm{ref}}-\mathrm{NRMSE}_{\mathrm{pol}}]_+\\
&-3.0[\mathrm{NRMSE}_{\mathrm{pol}}-\mathrm{NRMSE}_{\mathrm{ref}}]_+
-0.75[\mathrm{NRMSE}_{\mathrm{pol}}-\mathrm{NRMSE}_{\mathrm{best}}]_+,
\end{aligned}
\label{app:eq:reward}
\end{equation}
where $[z]_+=\max(z,0)$. The actor and critic objectives are
\begin{align}
\mathcal L_{\mathrm{actor}}
&=-\log\pi_\theta(a\mid s)\,\hat A
-\beta\mathcal H(\pi_\theta(\cdot\mid s))
+\lambda_{\mathrm{BC}}\mathcal L_{\mathrm{BC}},\\
\mathcal L_{\mathrm{critic}}
&=(V_\phi(s)-r)^2,
\end{align}
with one-step advantage $\hat A=r-V_\phi(s)$. Hyperparameters are listed in Table~\ref{app:tab:hyper}.

\begin{table}[htbp]
\centering
\caption{Reward components and their purposes.}
\label{tab:reward_components}
\scriptsize
\setlength{\tabcolsep}{3.4pt}
\begin{tabular}{lll}
\toprule
\textbf{Reward Component} &
\textbf{Formulation} &
\textbf{Purpose} \\
\midrule
Negative validation &
$-\mathrm{NRMSE}_{pol}$ &
Penalizes prediction error \\

Improvement bonus &
$+1.5[\mathrm{NRMSE}_{ref}-\mathrm{NRMSE}_{pol}]_+$ &
Rewards improvement over reference \\

Underperformance penalty &
$-3.0[\mathrm{NRMSE}_{pol}-\mathrm{NRMSE}_{ref}]_+$ &
Penalizes worse performance than reference \\

Expert-gap penalty &
$-0.75[\mathrm{NRMSE}_{pol}-\mathrm{NRMSE}_{best}]_+$ &
Penalizes worse performance than the validation-optimal mixture \\
\bottomrule
\end{tabular}
\end{table}

\subsection{Validation gate}

The deployed supervised fallback $g_{\mathrm{fb}}$ is the better of target-only scratch and all sources plus target on the target validation split. For policy predictor $g_{\mathrm{pol}}$,
\begin{equation}
g_{\mathrm{final}}=
\begin{cases}
g_{\mathrm{pol}}, &
\mathrm{NRMSE}_{\mathrm{pol,val}}
\le \mathrm{NRMSE}_{\mathrm{fb,val}}-\delta,\\
g_{\mathrm{fb}}, & \text{otherwise}.
\end{cases}
\label{app:eq:gate}
\end{equation}
The experiments use $\delta=0$. The raw actor--critic variant omits Eq.~\ref{app:eq:gate}; the behaviour-cloning variant omits actor--critic refinement but retains the same gate.

\section{Extended comparison and diagnostic results}
\label{app:results}

\begin{table}[htbp]
\centering
\caption{Overall performance across held-out target materials, fitting-data
budgets, and subset draws. RMSE is in $\mathrm{kgCO_2e/part}$; values are mean
$\pm$ standard deviation.}
\label{app:tab:overall}
\scriptsize
\setlength{\tabcolsep}{3.4pt}
\begin{tabular}{@{}lcccc@{}}
\toprule
Method & Val. $R^2$ & Val. RMSE & Test $R^2$ & Test RMSE \\
\midrule
Actor--critic + gate & $0.8580\pm0.1496$ & $0.0044\pm0.0015$ & $0.7757\pm0.2151$ & $0.0052\pm0.0021$ \\
Behaviour cloning + gate & $0.8303\pm0.2777$ & $0.0047\pm0.0020$ & $0.7264\pm0.2130$ & $0.0053\pm0.0022$ \\
Raw actor--critic & $0.8064\pm0.2535$ & $0.0053\pm0.0020$ & $0.7195\pm0.1893$ & $0.0056\pm0.0022$ \\
All sources + target & $0.8098\pm0.2679$ & $0.0050\pm0.0020$ & $0.6633\pm0.2654$ & $0.0058\pm0.0023$ \\
Target only & $0.6703\pm0.3684$ & $0.0066\pm0.0033$ & $0.4929\pm0.6609$ & $0.0065\pm0.0033$ \\
Adaptation simplex & $0.4851\pm0.6410$ & $0.0077\pm0.0036$ & $0.5005\pm0.3311$ & $0.0077\pm0.0039$ \\
Best single source & $0.4648\pm0.6292$ & $0.0082\pm0.0036$ & $0.4678\pm0.2988$ & $0.0082\pm0.0037$ \\
Pooled pretrain--finetune & $0.5170\pm0.5697$ & $0.0085\pm0.0040$ & $0.0696\pm1.1694$ & $0.0094\pm0.0049$ \\
Similarity-weighted & $-0.2340\pm1.3638$ & $0.0119\pm0.0049$ & $0.0921\pm0.6136$ & $0.0102\pm0.0055$ \\
Uniform average & $-0.2086\pm1.4216$ & $0.0114\pm0.0053$ & $-0.0095\pm0.7246$ & $0.0103\pm0.0059$ \\
Nearest-source pretrain--finetune & $-35.9234\pm116.1299$ & $0.0530\pm0.0788$ & $-64.6478\pm223.8527$ & $0.0528\pm0.0769$ \\
\bottomrule
\end{tabular}
\end{table}

The full benchmark places the policy against prediction-, data-, and
parameter-transfer alternatives. Simple averaging and similarity weighting do
not resolve incompatible sources, while the learned variants benefit from
case-specific candidate evidence. The behaviour-cloning and raw actor--critic
rows act as complementary ablations: together with the gated result, they show
how validation-oriented initialization, reward-based refinement, and supervised
fallback contribute to the deployed method.

\subsection{Budget-dependent performance}

\begin{figure}[H]
  \centering
  \includegraphics[width=\linewidth]{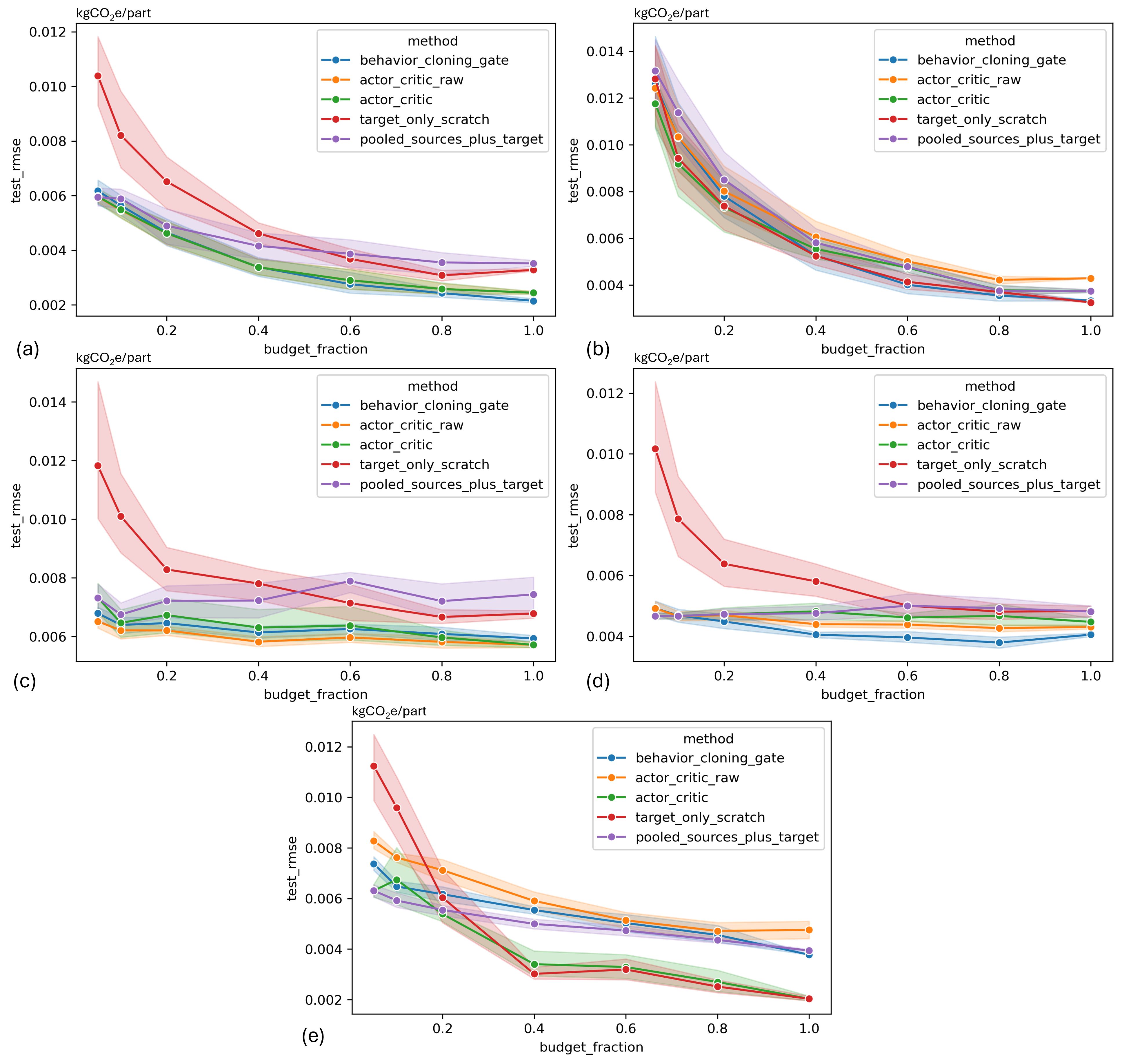}
  \caption{Test RMSE versus target fitting budget for (a) PETG, (b) ABS, (c) PLA, (d) rPLA, and (e) TPU. The budget $b$ is the fraction of the target training partition used for fitting; the separate validation partition remains fixed. Curves compare the complete policy, its ablations, target-only fitting, and all-sources-plus-target pooling. RMSE is expressed in $\mathrm{kgCO_2e/part}$. Lines show mean test RMSE; shaded bands show standard deviation.}
  \label{app:fig:budgets}
\end{figure}

No method dominates for every target and budget. Transfer-aware methods are most useful when target fitting data are scarce, but target-only fitting can become preferable as the target relationship becomes estimable. PETG shows a large low-budget gap: target-only RMSE is approximately 0.010 at $b=0.05$, while transfer-aware methods are near 0.006. ABS methods converge at larger budgets, consistent with weak source-expert transfer into ABS. TPU reverses from poor target-only performance at 5--10\% to target-only superiority after approximately 40\%. These results motivate conditioning the decision on both domain evidence and target fitting-data budget.

\subsection{Distance, negative transfer, and failure cases}

\begin{figure}[htbp]
  \centering
  \includegraphics[width=0.86\linewidth]{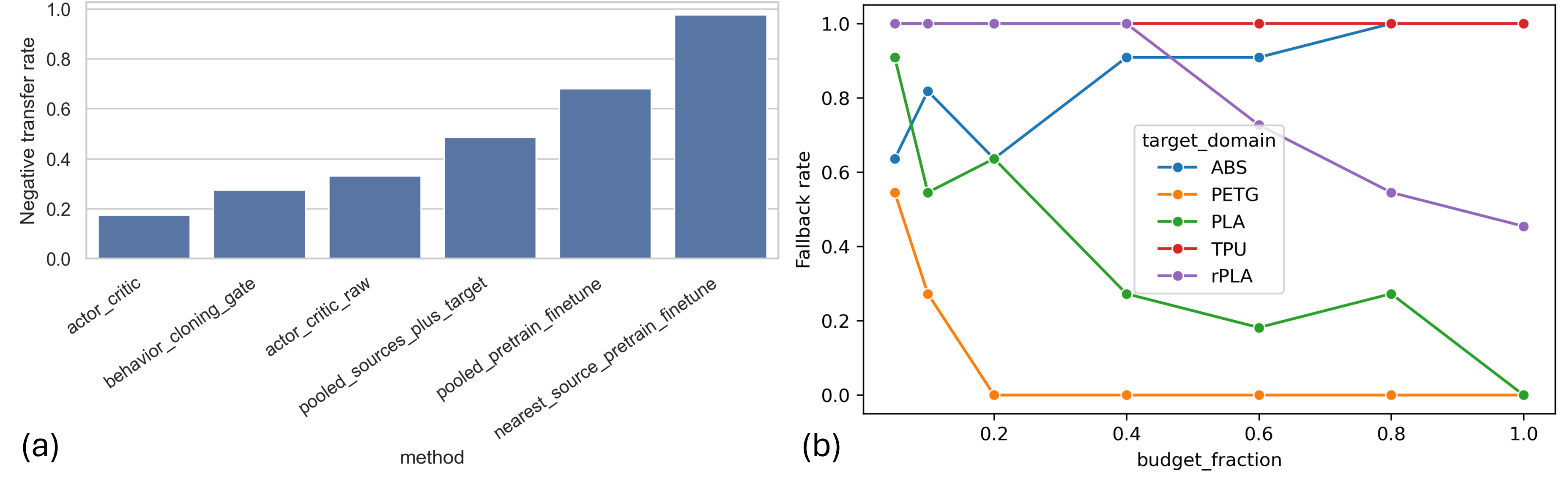}
  \caption{Deployment evidence: (a) negative-transfer rate relative to target-only fitting; and (b) frequency with which the validation gate replaces the actor--critic prediction with the stronger supervised model.}
  \label{app:fig:deployment}
\end{figure}

\begin{figure}[H]
  \centering
  \includegraphics[width=\linewidth]{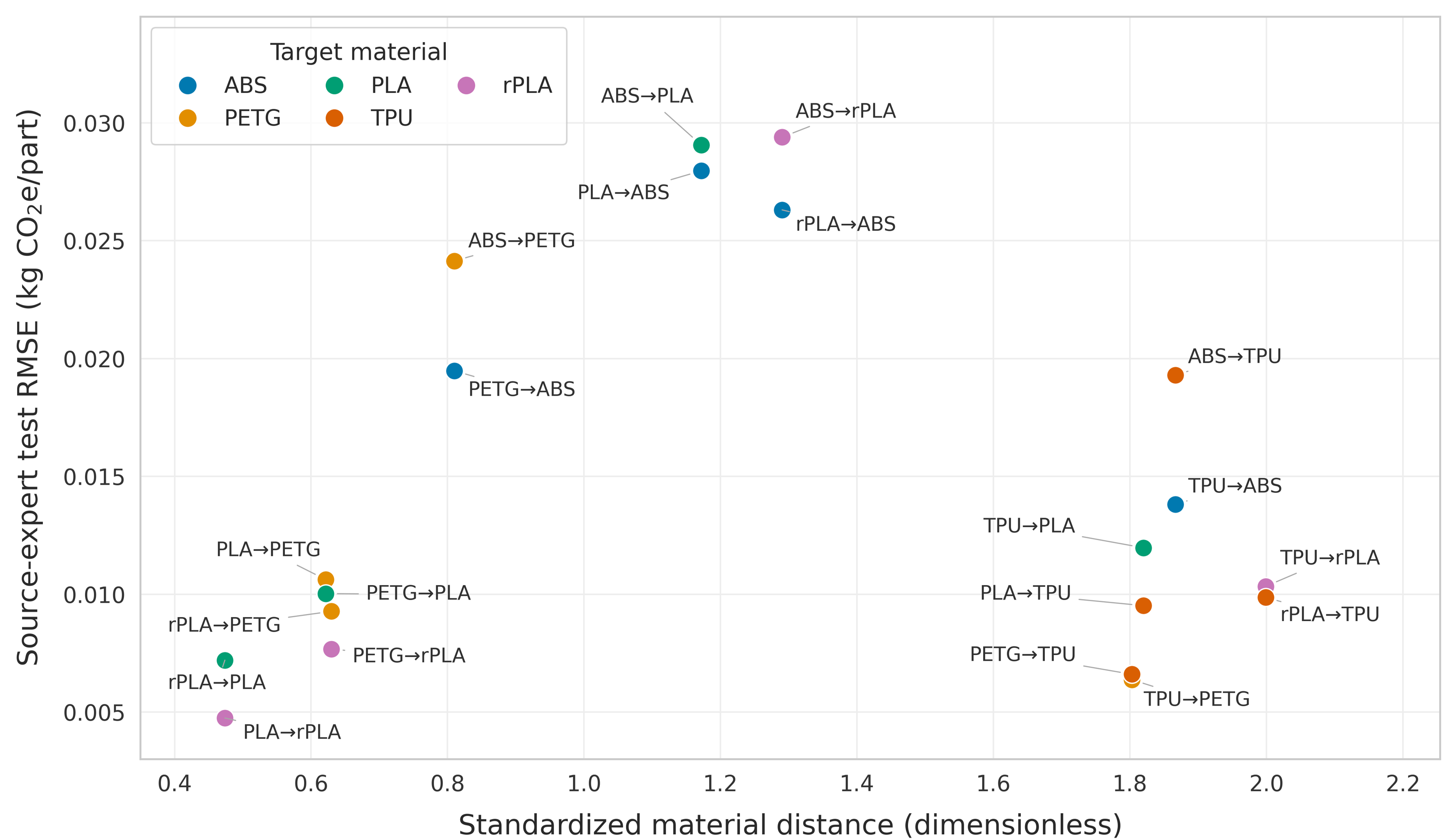}
  \caption{Source-expert test RMSE against standardized material distance for
  representative source--target pairs; statistics over all 20 directed pairs are given in Appendix~\ref{app:decision}.}
  \label{app:fig:distance}
\end{figure}

Material proximity is not a reliable stand-alone transfer rule. ABS--PETG has a standardized distance of 0.810 but poor transfer error near 0.019--0.024~$\mathrm{kgCO_2e/part}$, whereas PETG--TPU has a larger distance of 1.803 and transfers well in both directions. Distance is therefore retained as one state component rather than used as a deterministic decision.

The negative-transfer rate is approximately 0.17 for actor--critic plus gate, 0.28 for behaviour cloning plus gate, 0.33 for raw actor--critic, 0.49 for all sources plus target, 0.68 for pooled pretrain--finetune, and 0.98 for nearest-source pretrain--finetune. Thus, the gate reduces the raw policy's observed harmful-transfer rate by about 48\%. Fallback frequency is itself a result: ABS increasingly uses a supervised reference and reaches a fallback rate of 1.0 by $b=0.80$; TPU uses fallback in all cases at every budget. More target data can therefore make transfer less appropriate when the target-specific supervised relationship becomes stronger.

\begin{figure}[htbp]
  \centering
  \includegraphics[width=\linewidth]{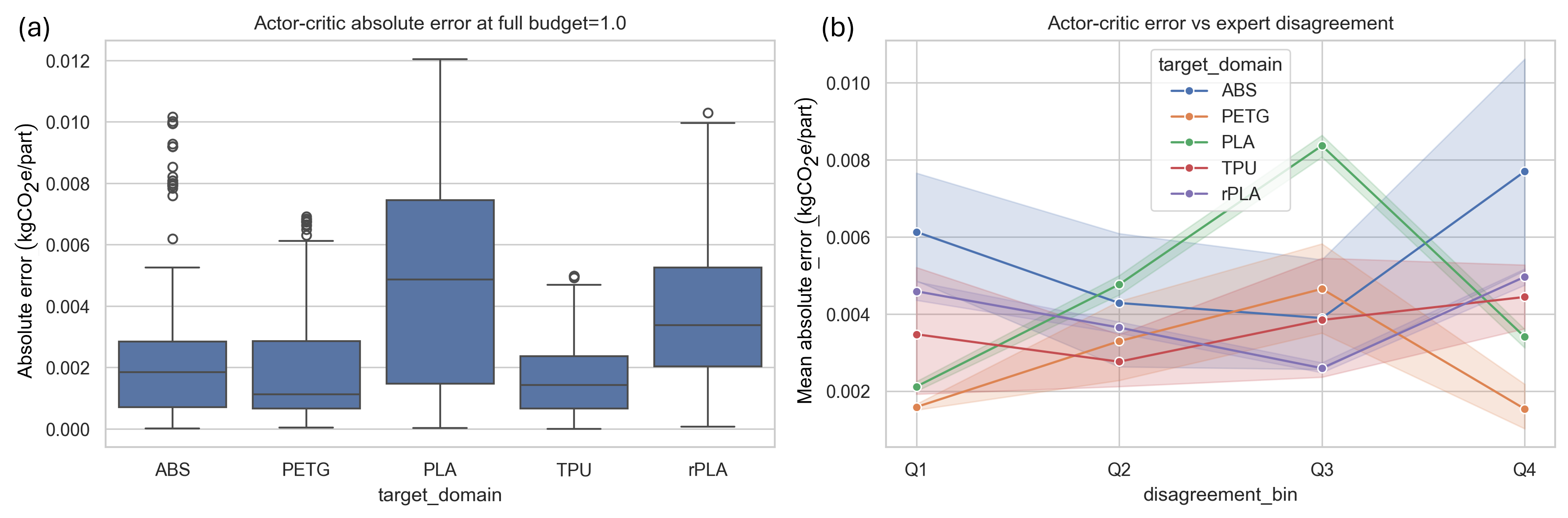}
  \caption{Failure analysis using (a) absolute prediction error and (b) mean absolute error by source-expert disagreement group.}
  \label{app:fig:failure}
\end{figure}

Residual error is target-specific and often coincides with disagreement among source experts. PLA shows the clearest sensitivity: mean absolute error rises from approximately $2.1\times10^{-3}$ in the first disagreement quantile to $8.4\times10^{-3}$ in the third. ABS also deteriorates in the highest quantile. This pattern supports the inclusion of expert disagreement in the policy state and identifies high-disagreement cases as priorities for additional target measurements.

\subsection{Manufacturing interpretation}

For implementation, a manufacturer needs a common predictor schema across historical and new operations, a small target fitting set, and a separate target validation set. The candidate bank then turns historical operations into auditable alternatives: individual experts expose directional compatibility, subset candidates capture complementary sources, and the target-only and pooled candidates anchor deployment. The selected predictor supplies an inventory-linked response before a full characterization campaign is complete, while disagreement and fallback frequency indicate where additional target measurements are most valuable.

The controlled cross-material setting holds geometry, printer, and response construction consistent, so the observed differences can be attributed to the transfer decision rather than changes in system boundaries. The demonstrated benefit is improved prediction of an operational-plus-material carbon indicator; realized emissions reductions depend on using those predictions to choose lower-footprint, technically feasible alternatives.

\section{Decision formulation, comparison with direct learning, and policy specification}
\label{app:decision}

\subsection{Prediction versus decision}
Direct supervised learning fits one predictor by minimizing prediction error on the data it is given; whether reusing source data helps or harms a new target becomes visible only when alternatives are compared. The proposed method learns that comparison as a decision. Its action selects and weights candidates, including target-only learning; its reward (Eq.~\ref{app:eq:reward}) scores the resulting mixture against the stronger supervised reference; and the gate (Eq.~\ref{app:eq:gate}) deploys the policy only when validation evidence favours it, using a validation-based comparison with a supervised fallback~\cite{thomas2015high,laroche2019safe}. 

Validation-optimal mixture weights provide behaviour-cloning targets on policy-training materials, followed by reward-based refinement; the trained policy is then evaluated on a held-out material. Each case has one state, one action, and one reward, so the problem is a contextual bandit, and one-step actor--critic, i.e.\ REINFORCE with a learned baseline~\cite{williams1992simple}, is the standard estimator. The score-function estimator requires only reward evaluations, which keeps the candidate bank, reward, and gate interchangeable. The Dirichlet distribution is the natural policy class on the simplex of mixture weights, as Beta policies are for bounded actions~\cite{chou2017improving}.

\textbf{Reward weights.} The weights in Eq.~\ref{app:eq:reward} encode the deployment preference. Underperforming the trusted supervised reference costs twice the credit given for improving on it (3.0 versus 1.5), and distance from the validation-optimal mixture is a weaker shaping term (0.75). They were set from this preference rather than tuned on held-out data, so the test results carry no weight-selection bias. The coefficients change the reward's relative emphasis on improvement and underperformance during policy training. Other processes, response scales, or risk tolerances may call for recalibration, ideally on validation data, and a systematic study of this choice is left to future work.

\textbf{Direct optimization.} When labelled target validation data exist, mixture weights can also be optimized directly on the target validation set (Eq.~\ref{app:eq:teacher}). We use this solution as the imitation target rather than as a deployed predictor. The learned policy instead provides one decision rule that is shared across materials, evaluated on materials it has not seen, and combined with a gate that guards against negative transfer. Direct differentiation through the mixture reward is also a possible approach to training a shared policy; the present experiments do not establish an advantage of the score-function estimator over this alternative. Comparisons with this approach and with per-target direct optimization are left to future work.

\subsection{Comparison with direct learning}
Table~\ref{app:tab:paired} adds 95\% intervals and paired tests over the 385 matched held-out cases. All cases of a material share its test set, trained policy, and validation set; within a material--budget combination they differ only in the repetition. Case-level tests therefore treat repetitions as replicates: they assess whether a difference is consistent across repetitions for the evaluated materials. Because the repetitions at $b=1$ share their fitting data, their cases are near-replicates; excluding them, the reduction in negative transfer relative to behaviour cloning with the gate remains consistent (53 versus 23 discordant cases, McNemar $p=8\times10^{-4}$). Generalization to new materials is assessed by a hierarchical bootstrap that resamples materials and then cases; with five materials, these intervals are approximate. The proposed method lowers RMSE by 0.64~g per part (95\% interval 0.36--0.87~g) relative to all sources plus target and by 1.31~g (0.63--1.90~g) relative to target-only fitting. It lowers the negative-transfer rate by 31 percentage points (95\% interval: 18--44 percentage points) relative to pooling. The hierarchical bootstrap intervals support an average improvement across the evaluated materials, although inference beyond these materials remains limited by the five-material sample. Relative to behaviour cloning with the gate, the reduction in negative transfer is consistent across repetitions (Table~\ref{app:tab:paired}) and concentrated in TPU and ABS, the targets with the weakest source transfer; for PLA and rPLA, behaviour cloning with the gate has slightly lower rates, so this comparison holds across repetitions rather than across materials. 

The decision is also robust across conditions. In each of the 35 material--budget cells, the proposed method was compared with the better of the two fixed rules, target-only fitting and pooling, chosen in hindsight from test results. The proposed method is within 5\% of, or better than, that hindsight choice in 29 cells, never more than 15\% worse, and 3.2\% better on average. Behaviour cloning with the gate and raw actor--critic are up to 86\% and 134\% worse, respectively. The spread of the average row of Table~\ref{tab:results} reflects the conditions tested more than instability. A one-way analysis of variance of the 385 per-case test RMSE values of the proposed method, grouped by the 35 material--budget combinations, attributes 83\% of the variance to differences between combinations ($\eta^2$; material alone 27\%, budget alone 34\%, the remainder their interaction) and 17\% to variation across repetitions within a combination. Per-part responses average 0.051~$\mathrm{kgCO_2e}$ across the 729 prints, so the overall RMSE of the proposed method is about 10\% of a typical part's footprint, against 13\% for target-only fitting.

\begin{table}[!h]
\centering
\caption{Held-out results with 95\% intervals (385 cases; intervals resample cases within material--budget cells) and paired tests against the proposed method (Wilcoxon signed-rank for RMSE, McNemar for negative transfer). Intervals and $p$-values are case-level and treat repetitions as replicates for the evaluated materials.}
\label{app:tab:paired}
\scriptsize
\setlength{\tabcolsep}{4pt}
\begin{tabular}{@{}lcccc@{}}
\toprule
Method & RMSE ($\mathrm{kgCO_2e/part}$) & Neg. transfer (\%) & $p$ (RMSE) & $p$ (neg. transfer) \\
\midrule
Actor--critic + gate (proposed) & 0.0052 [0.0051, 0.0053] & 17 [14, 20] & -- & -- \\
Behaviour cloning + gate & 0.0053 [0.0053, 0.0054] & 28 [25, 30] & 0.13 & $4\times10^{-5}$ \\
Raw actor--critic & 0.0056 [0.0055, 0.0057] & 33 [30, 36] & $1\times10^{-9}$ & $2\times10^{-11}$ \\
All sources + target & 0.0058 [0.0057, 0.0059] & 49 [45, 52] & $4\times10^{-30}$ & $5\times10^{-35}$ \\
Target only & 0.0065 [0.0063, 0.0067] & -- & $3\times10^{-31}$ & -- \\
\bottomrule
\end{tabular}
\end{table}

\subsection{Policy specification}
For a held-out target, the bank contains 17 candidates: four single-source experts, eleven source subsets, target-only scratch, and all sources plus target. For a policy-training case, candidates involving the pseudo-target are unavailable, leaving nine. The state has 81 features: the budget and fitting-set size; means and standard deviations of the 22 predictors over the fitting subset; per-source material distances, fitting-subset errors, and availability masks; summaries of expert disagreement; validation errors of the two supervised references; and aggregate summaries of these quantities. Policy-training and held-out cases use separate fitting-subset draws. Across all 20 directed source--target pairs, material distance and source-expert error are not significantly rank-correlated (Spearman $\rho=0.28$, $p=0.24$ on target validation data; $\rho=0.21$, $p=0.38$ on test data), and validation and test errors agree closely ($\rho=0.87$). The distance finding therefore does not depend on test data.

\begin{table}[htbp]
\centering
\caption{Policy hyperparameters.}
\label{app:tab:hyper}
\scriptsize
\begin{tabular}{@{}llll@{}}
\toprule
Actor and critic & $2\times32$ MLP, dropout 0.1 & Optimizer & Adam, lr $5\times10^{-4}$, weight decay $10^{-4}$ \\
Behaviour-cloning epochs & 500 & Refinement epochs & 500 \\
Entropy weight $\beta$ & 0.1 & Critic loss weight & 0.5 \\
$\lambda_{\mathrm{BC}}$ & 0.3 & Gradient-norm clip & 1.0 \\
Dirichlet concentration floor & 0.1 & Gate margin $\delta$ & 0 \\
Reward weights (Eq.~\ref{app:eq:reward}) & 1.5, 3.0, 0.75 & & \\
\bottomrule
\end{tabular}
\end{table}

\end{document}